\documentclass[journal,transmag,twoside]{IEEEtran}

\usepackage{graphics} 
\usepackage{epsfig} 
\usepackage{mathptmx} 
\usepackage{times} 
\usepackage{amsmath} 
\usepackage{amssymb}  
\usepackage{kotex}
\usepackage{color}
\usepackage{multirow}
\usepackage{subfigure}
\usepackage{cases}
\usepackage{xcolor}
\usepackage[linesnumbered,ruled,vlined]{algorithm2e}

\SetCommentSty{mycommfont}
\SetKwComment{Comment}{/* }{ */}
\SetKwInput{KwData}{INPUT}
\SetKwInput{KwResult}{OUTPUT}

\usepackage[numbers]{natbib}
\bibpunct{[}{]}{,}{n}{}{;}
\usepackage[noadjust]{cite}  

\graphicspath{{./images/}}
\DeclareGraphicsExtensions{.pdf,.png,.jpg}

\definecolor{orange}{rgb}{1,0.2,0}
\definecolor{OliveGreen}{rgb}{0,0.6,0}

\newcommand{\yhk}[1]{{{#1}}}

\newenvironment{tightitemize}
{
	\begin{list}{$\bullet$}{%
			\setlength{\leftmargin}{8pt}
			\setlength{\topsep}{1pt}
			\setlength{\partopsep}{0pt}
			\setlength{\itemsep}{2pt}
			\setlength\labelwidth{5pt}}
		\ignorespaces}
	{\unskip\end{list}}

\newcounter{tecounter}
\newenvironment{tightenumerate}
{
	\begin{list}{\arabic{tecounter}\addtocounter{tecounter}{1}.}{%
			\setcounter{tecounter}{1}
			\setlength{\leftmargin}{12pt}
			\setlength{\topsep}{1pt}
			\setlength{\partopsep}{0pt}
			\setlength{\itemsep}{2pt}
			\setlength\labelwidth{7pt}}
		\ignorespaces}
	{\unskip\end{list}
}

\begin{document}
\title{
		A Wide-area, Low-latency, and Power-efficient 6-DoF Pose Tracking System for Rigid Objects		
}

\author{Young-Ho Kim, Ankur Kapoor, Tommaso Mansi, and Ali Kamen 
\thanks{Y.-H. Kim, A. Kapoor, T. Mansi, and A. Kamen are with Siemens Healthineers, Digital Technology \& Innovation, Princeton, NJ, USA
	{\tt\small\{young-ho.kim, ankur.kapoor, tommaso.mansi, ali.kamen\}@siemens-healthineers.com}}%
\thanks{Y.-H. Kim is a corresponding author.}
}	

\IEEEtitleabstractindextext{%

\begin{abstract}

Position sensitive detectors (PSDs) offer possibility to track single active marker's two (or three) degrees of freedom (DoF) position with a high accuracy, while having a fast response time with high update frequency and low latency, all using a very simple signal processing circuit. 
However they are not particularly suitable for 6-DoF object pose tracking system due to lack of orientation measurement, limited tracking range, and sensitivity to environmental variation. 
We propose a novel 6-DoF pose tracking system for a rigid object tracking requiring a single active marker. The proposed system uses a stereo-based PSD pair and multiple Inertial Measurement Units (IMUs). This is done based on a practical approach to identify and control the power of Infrared-Light Emitting Diode (IR-LED) active markers, with an aim to increase the tracking work space and reduce the power consumption.
Our proposed tracking system is validated with three different work space sizes and for static and dynamic positional accuracy using robotic arm manipulator with three different dynamic motion patterns. 
\yhk{Results demonstrate that the static position accuracy root-mean-square (RMS) error is $0.6~mm$ ($0.22~mm$ precision RMS error) at the mid range of the workspace and $1.3~mm$ ($0.44~mm$ precision RMS error) at the largest workspace. The dynamic position RMS error is $0.7$-$0.9~mm$. The orientation RMS error is between $0.04^\circ$ and $0.9^\circ$ at varied dynamic motion.}
Overall, our proposed tracking system is capable of tracking a rigid object pose with sub-millimeter accuracy at the mid range of the work space and sub-degree accuracy for all work space under a lab setting.

\end{abstract}

}

\maketitle
\section{Introduction} \label{sec:introduction}

\IEEEPARstart{P}{osition} sensitive detectors (PSDs)\yhk{\,\citep{wallmark57psd,ludwig65psd}} have become one of the most important components in the 2 or 3 degree-of-freedom (DoF) position tracking system for various applications \citep{lee10psd,ivan12psd2d,qu19psd,yang17psd} ({\em e.g.,} primarily based on lasers for measuring distance, displacement, and vibration), for which they provide high position resolution, fast response time, while being rather cost-effective and requiring simple signal conditioning circuits. 



\yhk{PSD provides a continuous position measurement of the incident light spot on a surface featuring a special monolithic PIN (positive-intrinsic-negative) photodiode with several electrodes placed along each side of the square near the sensor boundary\,\citep{wallmark57psd,hamamatsu,ludwig65psd}. The currents collected from the electrodes directly provide the incident light position with a simple circuitry, which has nanoscale position and time resolutions leading to a fast response time and a high accuracy position tracking system.
	\citet{lee10psd} used two PSD sensors to track 3-DoF position as a stereo vision system.
	PSD is utilized for visual servoing and  control applications \citep{ivan12psd2d}. \citet{yang17psd} used PSD for measuring a string vibration primarily due its fast response time.
	\citet{qu19psd} applied PSD-based position detection system for a closed-loop control of a solar tracking mobile robot.
	There are also work done in literature to better characterize the performance of PSD based systems. 
	\citet{rodriguez16calibrationLPS} proposed a mathematical model and a calibration method to do accurate measurements using PSDs. PSD errors are analyzed in terms of component tolerances, temperature variations, signal to noise ratio, operational amplifier parameters, and analog to digital converter quantization\,\citep{rodriguez16analysis}. In addition, \citet{Lu19spot} presented a quantitative analysis on the position error caused by the changes in the light spot diameter and the distance.}

Despite advantages, PSDs are not generally used for 6-DoF object pose tracking system (position and orientation). This is primarily based on three reasons; 1) a single marker orientation can not be captured using a PSD alone, 2) a multi-marker recognition is required for orientation estimation and that is not readily feasible due to multiple markers' line-of-sight and extra computation to identify markers and compute the pose, and 3) sensitivity of positional accuracy to active light source usually a light emitting diode (LED) to interference such as ambient light. 

\begin{figure*}[t!]
	\begin{center}
		\includegraphics[scale= 0.40]{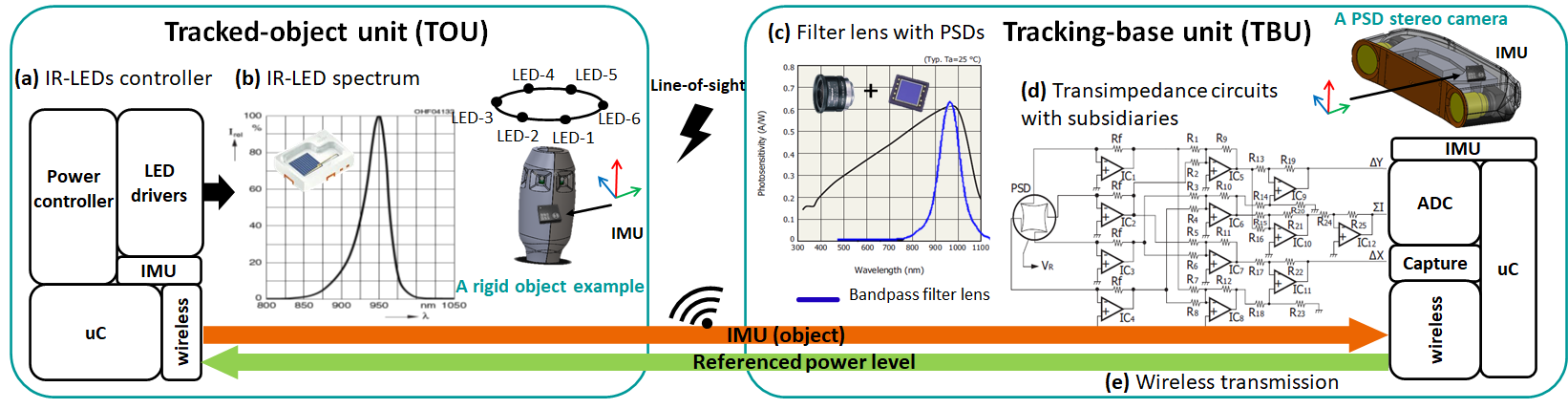}
		\caption{The proposed system consists of the tracking-base unit (TBU) and the tracked-object unit (TOU). (a)(b) TOU consists of multiple IR-LEDs (Centroid wavelength is 940nM.) along with one IMU. The IMU of TOU orientation information is transferred to the TBU via a wireless connection. (c)(d) TBU has two PSDs as a pair with a bandpass filter and transimpedance circuits. IR-LEDs incident light positions on each PSD are digitized by an ADC and processed to micro-controller via an analog signal processing circuit. 
			\label{fig:overall}}
		\vspace*{-20pt}
	\end{center}
\end{figure*}

\yhk{There exist three main categories of contact-less technologies to perform 6-DoF rigid object pose tracking; 1) Electromagnetic tracking system (ETS), 2) Optical-based  tracking system (OTS), 3) Other sensor fusions.}

Electromagnetic Tracking System (ETS) is mainly comprised of a stationary magnetic field generator, coil sensors attached to tracked objects, and a control unit. The magnetic field generator generates a magnetic field and establish a reference coordinate system. The coil sensors are attached to the tracked object, and coils induce a voltage due to magnetic field effect. The control unit operates the magnetic field generator, infers 6-DoF pose information from the coil sensors' voltages, and transfers the information to the host system\,\citep{chen20tracking}.  
ETS has a small size of coil sensors ({\em e.g.,} 1~mm in diameter and less than 10~mm in length), and does not require line-of-sight clearance. However, the tethered connection between the coil sensors and the control unit is cumbersome, and ETS accuracy is adversely affected by the presence of ferromagnetic object within the magnetic field of the field generator. Moreover, the tracking distance is limited up to $30~cm$, and the performance depends on the distance from the field generator\,\citep{franz14EM,andria20EM}. 

There exist two typical Optical Tracking Systems (OTS): Infrared (IR)-based and video-metric-based tracking system.
Depending on the type of fiducial markers used, IR-based tracking system can be categorized as either passive ({\em i.e.} using retro-reflective material) or active ({\em i.e.} IR emitting). The principle of tracking is based on triangulation and registration of a marker cluster, which is in fact a set of fiducial markers within a known geometry fixed to the tracked object. 3-DoF locations of each marker are estimated via triangulation where 3-DoF orientations are determined via registration to the known marker cluster geometry, requiring a high resolution/low latency cameras and non-trivial computations\,\citep{chen20tracking}. 

The number and spatial distribution of the fiducial markers significantly influence the tracking accuracy of the OTS. At least three visible and non-collinear markers are required to uniquely determine 6-DoF pose of the object. Moreover, the lighting conditions including natural environmental disturbance ({\em e.g.,} background illumination) may influence the tracking performance. 
The fiducial markers forming a marker cluster are usually  spatially distributed to provide a large lever-arm to provide rotational accuracy. Marker cluster size and the direct line-of-sight clearance are among disadvantages of OTS\,\citep{glossop09optical,sorriento20opticalEM}.

Video-metric-based tracking system is primarily relying on feature detection and image registration techniques for tracking objects. For example, a wearable 6-DoF hand pose tracking system is proposed for virtual reality application using a blob detection and tracking image processing algorithms for 3-DoF position tracking and IMU for 3-DoF orientation\,\citep{andualem17vr}.
\citet{garon17deep} introduced a temporal 6-DoF pose tracking system using deep learning algorithms with data augmentation. \citet{deng21posetracking} used a particle filter with deep learning methods to estimate 6-DoF targeted objects from camera images. In another example, \citet{dong20tracking} demonstrated data-driven methods to estimate a tracked object pose using a convolution neural network. These methods do not require markers, however training samples from the tracked object are needed to properly train the pose inference model. Moreover, the tracking accuracy is still limited as compared to OTS and ETS, which are capable of tracking the objects' pose with sub-millimeter and sub-degree accuracy with low latency ($30-80~Hz$) under an ideal setting.

Additionally, there exist multiple sensor-fusion-based 6-DoF pose tracking systems proposed in the literature. For example, \citet{han10ar} proposed a mobile robot pose tracking system for an augmented reality application, which integrated pose information from a radiofrequency-based pose tracking system with that from a vision-based tracking system. Electromagnetic tracking (5-DoF) and IMU (3-DoF) tracking data are fused to arrive at a full 6-D pose tracking resulted in\,\citep{dai18magnet}. \citet{esslinger20optoIMU} proposed a pose tracking system based on an opto-acoustic system and IMU, in which the fusing is done using a particle filtering approach.

\yhk{In this paper, we propose a new 6-DoF pose tracking system, enabling us to accurately track low power active markers within a large work space with high update rate and low latency. More specifically, the system has the following characteristics and novelties: 
\begin{tightitemize}
	\item The proposed system comprises of two main components; (1)~a tracking-base unit (TBU), housing two PSDs arranged with a fixed baseline to allow for position triangulation, and an inertial measurement unit (IMU) for a reference coordinate system; (2)~a compact size of tracked-object unit (TOU), which could be a set of active makers consisting of multiple infrared light emitting diodes (IR-LEDs) along with an inertia measurement unit (IMU). 
	\item A practical IR-LED identification methodology is proposed for a low-latency tracking and a noise filtering. This is primarily due to the limitation of PSDs in doing position sensing only for one active marker at a time and vulnerability to external IR ambient noise.
	\item An adaptable incident power control method is introduced to ensure optimal power consumption for the active markers and to increase the tracking range. 
	\item An overall calibration method using a highly accurate positioning system such as a commercial robotic manipulator is provided to properly fuse the rotational and translation information reads from the pair PSDs and IMUs.
    \item We finally integrate our tracking system with an ultrasound machine in order to track ultrasound probes, and demonstrate pose tracking performance in a lab environment.
\end{tightitemize}
}






\section{Materials and Methods}

%
%

Our objectives for the proposed system is to achieve a) high accuracy over a large work-space, b) high update rate along with low latency, c)  power efficiency to operate the sensors possibly with lithium ion batteries, and d) cost efficiency by utilizing readily available research and development cost amortized components. 

Our proposed system consists of two parts; {\it a tracking-base unit} (TBU) and {\it a tracked-object unit} (TOU) shown in Figure\,\ref{fig:overall}. TOU could have multiple light sources ({\em i.e.} active LEDs) along with an inertial measurement unit (IMU). 
The objective here is to have TOU as small and as low power as possible to have it running on battery. Furthermore, we included a micro-controller unit with wireless capabilities to transmit orientation parameters and also receive timing information and other commands from TBU in real-time.

TBU establishes the reference coordinate system and tracks one or more light spots emitted from the TOU using a pair of PSDs through triangulation. In TBU, we have one IMU to provide reference orientation information for measuring relative orientation as reported wirelessly from TOU. Specifically, the unique combination of each IR-LED and IMU ensures that position is measured by the stereo PSD system in TBU, while orientation is measured by the IMU of TOU in reference to the IMU in TBU. These two IMU-based orientation are coupled and provide a reference coordinate system at TBU for all measurements. 


We made three design decisions as follows:
\begin{tightenumerate}
	\item [{\bf 1. [3-DoF position]:}] Two PSD sensors are adopted for a stereo vision system instead of two CCD camera modules requiring imaging processing. PSDs are able to accurately detect the location of light spots from sensors, and two 2D detector locations can be triangulated into three-dimensional position in the stereo camera reference coordinate system.	
	\item [{\bf 2. [3-DoF orientation]:}] An estimated active IR-LED position is combined with orientation measurement from two IMUs in TOU and TBU; The relative orientation in the reference coordinate system is established by TBU's IMU. A rigid object's orientation is measured by IMU in TOU.
	\item[{\bf 3. [Multiple active markers]:}] In order to maintain a required line of sight and to track an object rotating a full $360$ degree around its axis, we need to be able to track multiple markers. Since the IR-LED's emitting angle is limited, we could instrument the tracked objects with a number of IR-LEDs to cover the rotational range. We need at least one IR-LED in the line-of-sight for each angle for TBU to detect.
\end{tightenumerate}

The TBU details are as follows: (1) Two PSD sensors (S5991, active area $9~mm \times 9~mm$, Hamamatsu) with optical camera lens (Forcal length $8.5~mm$, Angle of View (Horizonal $57.4^\circ$ ,Vertical $43.8^\circ$ ) and the band-pass filter ($940~nM$ center) are used as a stereo PSD system. (2) Analog signal processing circuits consisting of transimpedance, summation, addition, subtraction amplifiers for four PSD current signals (i.e., eight current signals from two PSD), described in Figure\,\ref{fig:overall}(d). We chose a rail-to-rail,  low-noise  OP-amp with low input bias current, such as OPA192 and OPA657. We used $1M\Omega$ as a feedback resistor, to achieve a high sensitivity of PSD. Thus, the system requires a hardware filter for external infrared noise removal. Analog-to-Digital (ADC) (AD7606, 6ch simultaneous sampling) circuit is employed to process the analog signal. Furthermore, we designed an event capture circuit to detect incident value without lagging. (3) IMU (BNO055, Bosch) is used to provide orientation measurements. (4) One micro-controller (CC2650, Texas Instrument) is used to gather and process signals, and send the results to the host computer in real-time. 

The TOU consist of four components; (1) Six IR-LEDs (SFH4775S, $940$~nM centroid wavelength, $120^\circ$ beam angle, Max $4.7$~V forward voltage, max $3$~A forward current at $100$~uS, Oslam), (2) N-Channel power MOSFETs (0.5A)  for each IR-LED,  (3) One IMU (BNO055, Bosch) rigidly attached to the IR-LED. (4) micro-controller and necessary circuit to feed LEDs with a pulse width modulated (PWM) variable constant amplitude voltage source. 


Each IMU reports an absolute orientation measurement with respect to the Earth and its magnetic field, which is obtained by fusing 9-DoF (accelerator, gyroscope, and magnetometer). BNO055 has internal fusion software, combining all three sensors with a fast calculation in high output data rate ($100$~Hz) and high robustness from other than the Earth magnetic field distortions.

We need to address the following four specific challenges for the proposed system.
\begin{tightenumerate}	
	 \item [{\bf A. [Identification of Multiple IR-LEDs]}]
	 Our proposed system have multiple IR-LEDs surrounding TOU, which requires to be tracked at least one IR-LED. However, it is still a critical point to recognize which single IR-LED is emitted from TOU because tracked each position of IR-LEDs will be integrated with IMU information to estimate a 6-DoF pose.
	 One possible solution is to construct a closed loop control system to synchronize TOU LED firing and TBU reading with a proper hand-shaking mechanism implemented through either a wireless or a wired connection. However, this takes extra times due to a limited bandwidth of transmitted packets, which is a bottleneck for achieving low latency. Lastly, it is desirable not to have additional wiring between TOU and TBU.
	 
	 \item [{\bf B. [Cancellation of natural environmental disturbance]}] 
     PSD has a wide spectral response range between 320 to 1100~nm even though it has a peak sensitivity wavelength around 960~nM. Thus, the incident light spot always includes unknown natural environmental disturbance, which lead to inaccurate measurement of position.
    
	 \item [{\bf C. [Active light intensity controls:]}] 
	 IR-LEDs illumination excite a spot on the PSD's 2D sensor and as the result a transimpedance circuit converts the current to a voltage, which is digitalized through an ADC. For large work space tracking, there are three important elements, namely  {\it IR-LED emitted light intensity}, {\it PSD sensitivity level in detecting IR-LED illumination}, and finally {\it ADC input range}.
	 To track the incident light in a large workspace, the light intensity needs to be increased exponentially over distance, but this increases IR-LED power consumption and temperature. \yhk{PSD sensitivity level might need to be increased over distance, but this could lead to vulnerability to external IR ambient noise.} Finally, increased ADC input range results in decreased ADC resolution.
	 
	 \item [{\bf D. [Calibration with PSDs and IMUs:]}]
	 To achieve 6-DOF pose tracking system using PSDs with IMUs, there exist three calibration stages. (1) A stereo PSD calibration to compute the depth from two 2D PSDs' positions. (2) the IMU coordinate system of TOU needs to be aligned with TBU reference coordinate system.
	 (3) The stereo PSD system has the reference coordinate assumed to be at the center of the left PSD sensor (or the right), which should be calibrated with the IMU coordinate system in TBU. 
\end{tightenumerate}

\begin{figure}[t]
	\begin{center}
		\includegraphics[scale= 0.35]{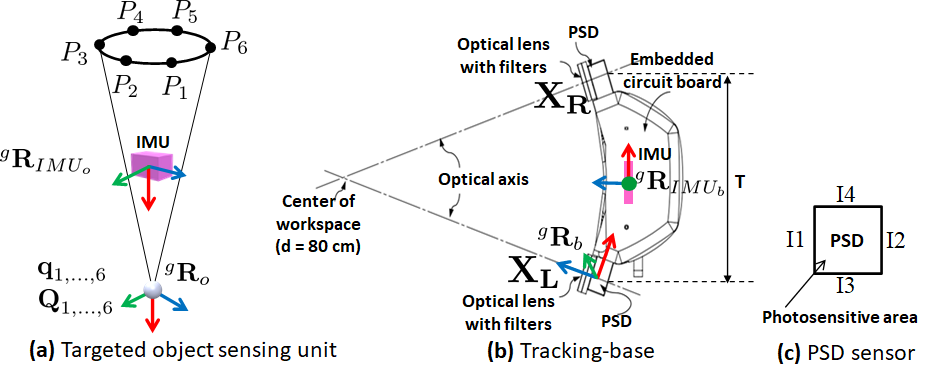}
		\caption{Nomenclature of proposed 6-DoF pose tracking system \label{fig:nomenclature}}
		\vspace*{-20pt}
	\end{center}
\end{figure}
 
The nomenclature of the proposed tracking system is described in Figure\,\ref{fig:nomenclature}. Let ${\bf P_i}$ represents $(x,y,z) \in \mathbb{R}^3$ as 3-DoF position of $i$-th IR-LED in terms of TBU's reference coordinate system ({\em i.e.} the left PSD coordinate system).
Two PSD's 2-DoF states ${\bf X_L} =(x_l,y_l)$, ${\bf X_R} =(x_r,y_r)$ has $(x,y) \in \mathbb{R}^2$ as 2-DoF positions of left and right PSD sensors, respectively. Let $I_i$ be a photocurrent from electrodes of PSD where $i \leq 4$ for four corners. ${\bf X_L}$ and ${\bf X_R}$ can be computed as follows;
\begin{eqnarray}\label{eq:psd}
{\small
\begin{aligned}
x &= \frac{L}{2}\frac{(I_2+I_3)-(I_1+I_4)}{(I_1+I_2+I_3+I_4)},~
y &= \frac{L}{2}\frac{(I_2+I_4)-(I_1+I_3)}{(I_1+I_2+I_3+I_4)}, 
\end{aligned}
}
\end{eqnarray}
where  $L$ is the resistance length of PSD, provided by the manufactures of PSD.

Let $^{g}{\bf R}_{IMU_o}$ and $^{g}{\bf R}_{IMU_b}$ be an absolute rotation matrix with respect to the earth ({\it g}) and its magnetic field, coming from the IMU in the {\it object, 'o'} TOU and the IMU in the {\it base, 'b'} TBU, respectively. Similarly, let $^{g}{\bf R}_{o}$ and $^{g}{\bf R}_{b}$ be the absolute orientation for TOU and TBU, respectively.

Let ${\bf q}_i$ be the relative pivoting 3-DoF position from each IR-LED $P_i$ position to the pivot point of the targeted object, while ${\bf Q}_i$ be the relative pivoting 3-DoF position with regard to the reference coordinate system in TBU ({\em i.e.,} the left PSD coordinate system).


\subsection{Identification of Multiple IR-LEDs}

There exist analog and digital multiplexing methodologies ({\em i.e.,} Frequency Division Multiplexing (FDM) and Time Division Multiplexing (TDM)) to track multiple active markers simultaneously. FDM is easy to synchronize multiple signals, however it requires complex circuitry to treat highly sensible multiple analog signals, and a large enough bandwidth channels to get highly accurate multiple signals without crosstalk. 
Instead, we bring a simple TDM idea with a trigger signal to recognize which IR-LED is excited.

We propose a pattern-based LED identification method between TBU and TOU, which is based on a pre-defined pattern used for each LED initiated from TOU. Thus, it does not require a hand-shake and the corresponding communication. 
We define a pattern signal period based on inverse of desired update frequency (e.g., $100$~Hz ). One pattern signal cycle consists of one generic trigger signal to indicate the start of the cycle and IR-LED specific signals as shown in Figure\,\ref{fig:identification}. 
More specifically for a 6-LED TOU, we have seven signals with even widths fitting in $10$~ms for $100$~Hz update frequency. The trigger signal at the start of pattern cycle is designed to be more than $50\%$ duty cycle Pulse Width Modulation (PWM) signal, exciting all IR-LEDs at the same time. Following that, IR-LED specific signals are designed to be less than $50\%$ duty cycle PWM signal, and is applied sequentially to IR-LEDs with a specific known consistent order. 

One exemplary full pattern signal is captured and depicted in Figure\,\ref{fig:identification}. TOU emits a pattern signal consisting of trigger and IR-LED specific patterns continuously. As the result, the emitted light from IR-LEDs pulses are detected by PSDs in the TBU through analog signal processing circuits (Purple in Figure\,\ref{fig:identification}). We devised an event capturing circuit to detect the rising edge of detected pulses (Green in Figure\,\ref{fig:identification}). 

We perform  two ADC conversions synced on the rising edge with a fixed interval (Cyan in Figure\,\ref{fig:identification}). After each ADC conversion, the results of ADC is read by the  micro-controller and used for further processing.
Then, the first trigger signal is easily detected in the TBU as two ADC conversion values are nearly the same and positive due to the larger Power and fixed reading interval (see first two rising edges (cyan) and the trigger signal (first pulse of purple) in Figure\,\ref{fig:identification}). Following this, as the IR-LEDs pattern follow a sequential order, the timing of the next non-zero ADC value determines the index of the exited IR-LED.

In the circular arrangement of IR-LEDS, only one or two LEDs are detected at a time within a single pattern cycle. This pattern-based approach provides a synchronous data processing in a real time without physical wire/wireless communication between TOU and TBU.

\subsection{Cancellation of natural environmental disturbance}\label{sec:cancel}
\begin{figure}[t]
	\begin{center}
		\includegraphics[scale= 0.63]{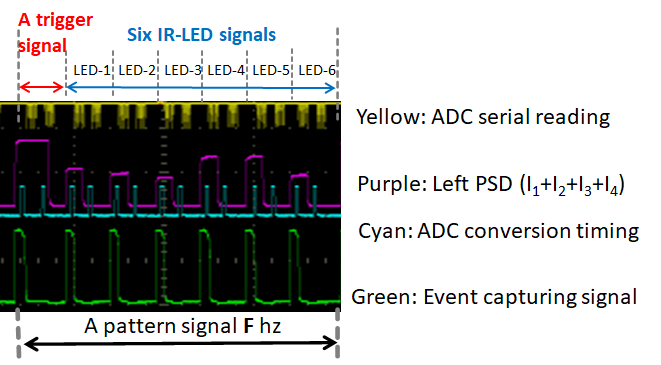}
		\vspace*{-10pt}
		\caption{Pattern-based IR-LEDs identification: We demonstrate one exemplary pattern signal (a trigger signal + multiple IR-LED signals) to explain the overall process. In reality, only one and two IR-LEDs are detected due to geometry constraints. \label{fig:identification}}
		\vspace*{-20pt}
	\end{center}
\end{figure}

Most tracking environment is not ideal settings including unknown environmental noises, which affects the performance of multiplexing method. \yhk{Thus, we propose two ADC converted values (cyan in Figure\,\ref{fig:identification}) for each signal ({\em i.e.} one trigger and multiple IR-LED signals).}

\yhk{Each signal is read by double ADC conversions.} The first one represents a IR-LED pure signal plus any possible ambient and other noises. The second ADC value captures external various noises primarily based on the ambient stray light, which is closely captured to the main signal captured in the first readout. Using these two, we have an opportunity to cancel out the background illumination by subtracting the second value from the first one. This is an important feature, which proved to be essential in reducing the effect of ambient stray light and other noises in an operating environment ({\em e.g.,} where the sun light or other sources of IR might be present). 
PSD output signal after circuitry has three analog signals; two subtractions (${I_2+I_3-I_1-I_4}$,  ${I_2+I_4-I_1-I_3}$), one summation (${I_1+I_2+I_3+I_4}$). We apply our proposed methodology for these three analog signals in a real-time ({\em i.e.,} 50~us capturing delay for each signal).

Overall, the advantages of this method are; (1) there is no time consuming hand-shake involving back and forth communication is required, (2) the external IR ambient noise can be alleviated, and finally (3) the centralized control system of IR-LEDs facilitates further the power adjustment and control for a large work-space, which we will be further explained in  the next section.


\subsection{Active light intensity controls}
IR-LED power needs to be continuously adjusted to provide better signal and overall accuracy specifically in scenarios where the distance between the TBU and TOU is large or steep orientation of TOU causes decreased signal from IR-LEDs. The basic idea is to construct a closed loop power controller and a look-up-table to adjust the power range depending on distance.
Given fixed PSD sensitivity level, and ADC range of inputs, the operational maximum power is a non-linear function of distance between the source and the detector. We construct the function as $target = LUT(d,R)$, where $d = { x_l}-{ x_r}$ is the sensor-based disparity (inversely proportional to the distance between the source and the detector), and $R$ is the orientation of TOU in terms of TBU coordinate system (detailed computation will be addressed in Section\,\ref{sec:calibration}).

\begin{algorithm}[b]
\caption{LIGHT$\_$INTENSITY$\_$CONTROLS~}\label{alg:LUT}
\KwData{the look-up-table for power limits LUT($\cdot$), the sum currents of left/right PSDs $\sum I_l$ and $\sum I_r$, x-axis positions of left/right PSDs $x_l$ and $x_r$ ($d = { x_l}-{ x_r}$), the orientation $R$}
\KwResult{$error$ }
$error = 0$\;
\While{PSD pose tracking system is OPERATIONAL}{
    ${target} = \text{LUT(d, R)}$ \label{step:lut}\;
    $error = target- \max(\sum I_l,\sum I_r)$ \label{step:errors}\;	  
}
\end{algorithm}


Active power controller algorithm is addressed in Algorithm\,\ref{alg:LUT}. The inputs are two summation values ($\sum I_l$, $\sum I_r$) from two PSDs, $x_l$, $x_r$, and $R$. During the operation, $LUT(\cdot)$ returns the reference power level. The discrepancy between the values denoted as error is  transferred to the TOU wirelessly allowing for the the power level of IR-LED to be controlled in real-time.

\subsection{Calibration Procedure \label{sec:calibration}}
We have mainly two calibration challenges; 1) PSD based stereo-camera calibration within in TBU, 2) 3-DoF position and 3-DoF orientation calibration within a unified coordinate system for TOU in terms of TBU.

\subsubsection{Stereo PSD calibration}

\begin{figure}[t]
	\begin{center}
		\includegraphics[scale= 0.4]{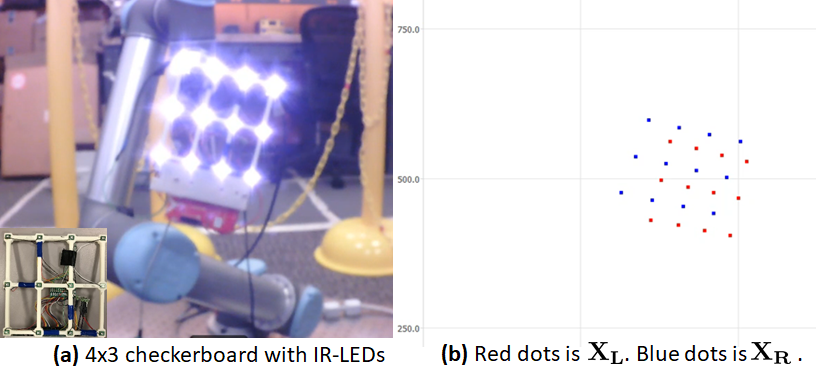}
		\vspace*{-10pt}
		\caption{Stereo PSD calibration for IR-LEDs: (a) 4x3 checkerboard is designed and used for stereo calibration. IR-band is not visible, so a single optical camera without IR-filter lens used to show how IR-LEDs are operated, which are emitted periodically. (b) 12 points detected by each PSD (left and right) are plotted in 2-DoF plane; the red is for ${\bf X_L}$, and the blue is for ${\bf X_R}$.
			 \label{fig:calibration}}
		\vspace*{-10pt}
	\end{center}
\end{figure}

Two PSDs are setup as a stereo camera system, where each PSD gives the 2D image point of LED point source, ${\bf X_L}$, ${\bf X_R}$ (Figure\,\ref{fig:calibration}(b)). Therefore similar to a stereo vision calibration, the PSD stereo calibration can be resolved using an intrinsic and extrinsic calibration steps. The intrinsic calibration parameters consist of the focal length, the principal point, skew angle and the ones related to distortion. Aside from distortion parameters, the focal length, skew angle, and the principal point  are directly analogous to those from commonly used cameras. 

To obtain projected points, we use a grid of IR-LEDs ({\em i.e.} to simulate an optical checkerboard) that is fabricated to certain tolerance to project $4 \times 3$ points in space (Figure\,\ref{fig:calibration}(a)). The checkerboard is moved around to cover the entire work-space area. Both the standard optical system parameters and the distortion coefficient are obtained using an iterative algorithm of projecting known points, de-warping and recasting them in the 3D space. Once the individual PSDs have been calibrated, we proceed with the calibration of extrinsic parameters of stereo rig, which is formulated as the relative  orientation and translation of the right PSD senor system with respect to the left PSD sensor. The extrinsic calibration parameters are computed using an iterative approach detailed in\,\citep{Bouguet01CameraCT}.

The distortion for PSD is different from CMOS/CCD type sensors used in computer vision. The typical biconvex lenses used to focus light produce barrel distortions, whereas the sensor itself produces a pincushion distortion. 

Ideally if the electrical center of PSD and the optical center of lens coincide the two can almost cancel out. However, this may not be achieved in practice. Therefore, we use a two dimensional Bernstein basis polynomials of degree $n$ to model the inverse of distortion. The formulation is as follows;
\begin{equation}\label{eq:distortion}
\begin{aligned}
(\hat{x},\hat{y}) &= \sum\sum \beta_{ij}b^n_{ij}(x,y)
\end{aligned}
\end{equation}

To obtain true 3D positions of IR-LED points, we apply the inverse of distortion modeled as in Equation\,\eqref{eq:distortion} to the PSDs measured pixels (${\bf X_L}$, ${\bf X_R}$) prior to 3D reconstruction of the point ($P_i$).

\subsubsection{Calibration of IMU and PSD coordinate system}

The 3-D position of the IR-LED source (i.e., attached to the tracked object) with respect to the left PSD coordinate system is given by ${\bf P}_i$. The 3-D orientation of the IMU in TOU is provided by $^g{\bf R}_{IMU_o}$ with respect to the gravity and the magnetic north. If the fixed relative orientation between the tracked object body and the attached IMU is given by $^{IMU_o}{\bf R}_{o}$, then the 3D orientation of the body with respect to gravity and magnetic north is follows:
\begin{equation}\label{eq:object}
\begin{aligned}
^{g}{\bf R}_{o}  = ^{g}{\bf R}_{IMU_o} \times ^{IMU_o}{\bf R}_{o}.
\end{aligned}
\end{equation}
 
Likewise, the relative orientation of the IMU attached to TBU is given by $^g{\bf R}_{IMU_b}$ and the fixed relative orientation between the left PSD coordinate system and the attached IMU on TBU is given by $^{IMU_b}{\bf R}_{b}$. Thus, the left PSD orientation (i.e., tracking base)  with respect to gravity and magnetic north is given by
\begin{equation}\label{eq:base}
\begin{aligned}
 ^g{\bf R}_{b} =  ^g{\bf R}_{IMU_b} \times ^{IMU_b}{\bf R}_{b}.
\end{aligned}
\end{equation}

Based on these, the 3-DoF orientation of tracked object unit with respect to left PSD coordinate system is given by 
\begin{equation}\label{eq:pose}
\begin{aligned}
^b{\bf R}_o &= ^g{\bf R}^{-1}_b \times ^g{\bf R}_o.
\end{aligned}
\end{equation}

Combination of $^b{\bf R}_o$ and ${\bf P}_i$, provide full 6 DoF transformation of the tracked object points to the tracking base. For example, any point in i-th IR-LED coordinate system such as the pivot point ${\bf q}_i$  has the following relationship to the same coordinate denoted in the tracking base coordinate system ${\bf Q}_i$. 
\begin{eqnarray}\label{eq:q}
{\bf Q}_i = ^{b}{\bf R}_{o} \times {\bf q}_i + {\bf P}_i.
\end{eqnarray}

Based on these, the three fixed quantities to be derived or calibrated for the PSD system are; {\bf (i)} relative orientation between tracked object IMU and the tracked object coordinate system  $^{IMU_o}{\bf R}_o$, {\bf (ii)} the relative orientation between the IMU and tracking base coordinate systems, $^{IMU_b}{\bf R}_b$,  {\bf (iii)} the relative position of a common point such as pivoting point denoted as ${\bf q_i}$ and ${\bf Q_i}$  within TOU coordinate system.

\vspace*{8pt}
\noindent{\bf Calibration of (i) $^{IMU_o}{\bf R}_{o}$:~}\\
$^{IMU_o}{\bf R}_{o}$ is stemming misalignment between IMU reference and the tracked object reference coordinate system. In a perfect scenario, this orientation transformation should be close to identity, but it is not in reality. To estimate $^{IMU_o}{\bf R}_{o}$, we have to measure the reference orientation of the tracked rigid object along TOU coordinate system, which might require the specialized tool ({\em e.g.,} a robotic manipulator). Let us have $M$ number of samples for a real measure of $^{g}{\bf \bar{R}}_{o}$, and TOU's pure IMU measure $^g{\bf R}_{IMU_o}$. \yhk{The real measure $^{g}{\bf \bar{R}}_{o}$ is directly obtained from the robotic manipulator ({\em i.e.} an end-effector pose information from kinematics). Then, from Equation\,\eqref{eq:object}, we can construct an overdetermined linear system as follows}
 
\begin{eqnarray}\label{eq:object_tip_equation}
(^{g}{\bf \bar{R}}_{o}^j-^{g}{\bf \bar{R}}_{o}^k)^{3P \times 3}  = (^{g}{\bf R}_{IMU_o}^j - ^{g}{\bf R}_{IMU_o}^k)^{3P \times 3} \times ^{IMU_o}{\bf \hat{R}}_{o},
\end{eqnarray}
where $j,k \in [1,M]$ and $M$ is the number of samples for $^{g}{\bf \bar{R}}_{o}$ and $^g{\bf R}_{IMU_o}$,  $P$ is the number of possible pair in $[1,M]$, $P >3$,  and $^{IMU_o}{\bf \hat{R}}_{o}$ is what we want to compute.

To minimize a residual error, we use the pseudo-inverse of $(^{g}{\bf R}_{IMU_o}^j - ^{g}{\bf R}_{IMU_o}^k)^{3P \times 3}$ to estimate $^{IMU_o}{\bf \hat{R}}_{o}$, then we can compute $^{g}{\bf \hat{R}}_{o}$. More detailed least-squares solution with pseudo-inverse is described in\,\citep{strang06linear}.

\yhk{Now we have $^{IMU_o}{\bf R}_{o}$, so we can compute $^{g}{\bf R}_{o}$. The next step is to estimate $^{IMU_b}{\bf R}_{b}$ to compute $^{g}{\bf R}_{b}$ (Equation\,\eqref{eq:pose}).}

\vspace*{8pt}
\noindent{\bf Calibration of (ii) $^{IMU_b}{\bf R}_{b}$ and (iii) ${\bf q_i}$:~}\\
We have to estimate $^{IMU_b}{\bf R}_{b}$, which is also mostly due to fabrication tolerance. Moreover, in our system, the relative position of each LED to the pivot point ${\bf q}_i$ must be determined in practice. We propose an extended pivot calibration method to estimate both the tolerance error $^{IMU_b}{\bf R}_{b}$ and each relative pivoting position ${\bf q_i}$ together.

The pivot calibration method evolves basically rotating and swinging the tracked object about a fixed point, while measuring the pose in both local (i.e., TOU) and global (i.e, TBU). The pivot point is constant within both local and global coordinate systems and that is the basis to create an over-complete set of equations.

In our system, each IR-LED needs individual calibration for ${\bf q_i}$, which is the location of each IR-LED is a coordinate system established at the pivot point. However, $^{IMU_b}{\bf R}_{b}$ is common for all IR-LED. We extend the pivot calibration method for each IR-LED 3-D position ${\bf P}_i$ to compute ${\bf q}_i$ where ${\bf Q}_1 = {\bf Q}_2 = \cdots = {\bf Q}_N$ is an ideal for all IR-LEDs.

We design an iterative least-square approximate method. 
Algorithm\,\ref{alg:iterative} consisted of three parts; (1) estimation of ${\bf q}_i$ and ${\bf Q}_i$ for $i$-th LED, (2) estimation of $^{IMU_b}{\bf R}_{b}$ for all LEDs, (3) computation of overall errors, and then iterate (1) to (3).

\yhk{First, we collect many 3-D points ${\bf P}_i$ for each $i$-th IR-LED associated with the orientation $^b{\bf \hat{R}}_o$. We assume that all collected data have a common {\bf Q} as the pivot point. To do that, a specialized tool like robotic manipulators might be helpful to collect data.
Then, we treat $\Delta_{^{IMU_b}\hat{\bf R}_{b}}$ as our estimated value for $^{IMU_b}\hat{\bf R}_{b}$, which is initially an identity matrix.
Let ${\bf S}_i^j = [{\bf P}^j, ^b{\bf \hat{R}}_o^j]$ denote $j$-th set of position and orientation for $i$-th LED, where $j \in [1,N_i]$, and $N_i$ is the total number of data set for $i$-th LED. 
Then, we can derive an overdetermined linear system for ${\bf q}_i$ and ${\bf Q}_i$ based on Equation\,\eqref{eq:q}.}

{\small
\begin{eqnarray}\label{eq:estimationRo_process1}
    {(^b{\bf \hat{R}}_o^p - ^b{\bf \hat{R}}_o^k)}^{3H \times 3} \times {\bf q}_i = ({\bf P}^k-{\bf P}^p)^{3H \times 1},\\
    {({^b{\bf \hat{R}}_o^p}^{-1} - {^b{\bf \hat{R}}_o^k}^{-1})}^{3H \times 3} \times {\bf Q}_i = ({^b{\bf \hat{R}}_o^p}^{-1}\cdot {\bf P}^p- {^b{\bf \hat{R}}_o^k}^{-1}\cdot {\bf P}^k)^{3H \times 1},\label{eq:estimationRo_process2}
\end{eqnarray}
}where $p,k \in [1,N_i]$, $H$ is the number of possible pair $\in [1,N_i]$.
Then, we use the pseudo-inverse to estimate ${\bf q}_i$ and ${\bf Q}_i$.

Second, We update $\Delta_{^{IMU_b}\hat{\bf R}_{b}}$ where ${\bf Q}_1 = {\bf Q}_2 = \cdots = {\bf Q}_N$. 
Given $^g{\bf \hat{R}}_o$, $^{g}{\bf R}_{IMU_b}$, the estimated $^b{\bf {R}}_o$ is defined as

{\small
\begin{equation}\label{eq:estimationRo}
    ^b{\bf \hat{R}}_o = {\Delta_{^{IMU_b}\hat{\bf R}_{b}}}^{-1} \times ^{g}{\bf R}_{IMU_b}^{-1} \times ^g{\bf \hat{R}}_o,
\end{equation}
}
where ${\Delta_{^{IMU_b}\hat{\bf R}_{b}}}$ is what we want to estimate. 

Then, we can construct another overdetermined linear system as follows;
{\small
\begin{equation}\label{eq:estimationRo2}
    {\Delta_{^{IMU_b}\hat{\bf R}_{b}}}^{-1} \times (^{g}{\bf R}_{IMU_b}^{-1} \times ^g{\bf \hat{R}}_o \times{\bf q}_i)^{3H \times 1} = ({\bf Q}_i - {\bf P}_i)^{3H \times 1}
\end{equation}
}
where $H$ is the number of possible pair $\in [1,N_i]$. Then, we also use the pseudo-inverse to estimate $^{IMU_b}\hat{\bf R}_{b}$ ({\em i.e.} $\Delta_{^{IMU_b}\hat{\bf R}_{b}}$)\,\footnote{$x'A = b'$  equal to $A'x = b$}.

Finally, we compute overall errors of our estimation and keep iterating three steps until threshold or max-iteration meet. As a result, we can get $\hat{\bf q}_i$,~$\hat{\bf Q}_i$, and $^{IMU_b}\hat{\bf R}_{b}$.


\begin{algorithm}[t]
\caption{ITEREATIVE$\_$METHOD~(${\bf S}_i^{[1,N_i]}$)}\label{alg:iterative}
\KwData{(${\bf S}_1^{[1,N_1]}, \dots,{\bf S}_i^{[1,N_i]}$): $N_i$ number of samples for i-th LED}
\KwResult{$\hat{\bf q}_i$,~$\hat{\bf Q}_i$,~$^{IMU_b}\hat{\bf R}_{b}$ }
$error = 0$, $^{IMU_b}\hat{\bf R}_{b} = \Delta_{^{IMU_b}\hat{\bf R}_{b}} = \mathbb{I}$ \tcp*[r]{Initialization}

\While{MAX-Iteration \& errors $>$ threshold}{
    \tcp{First: estimation of ${\bf q}_i$ and ${\bf Q}$}
    $RR=[], TT=[], errors = 0$\;
    \For{$i=1:6$}
    {
        $r=[]$, $t=[]$,$R=[]$, $T=[]$\;
       \For{$j=1:N_i-1$}{
           $^{b}{\bf \hat{R}}_{o}^{j} = \Delta_{^{IMU_b}\hat{\bf R}_{b}} \times ^{b}{\bf R}_{o}^{j}$\;
            \For{$k=1+j:N_i$}{
                $^{b}{\bf \hat{R}}_{o}^{k} = \Delta_{^{IMU_b}\hat{\bf R}_{b}} \times ^{b}{\bf R}_{o}^{k}$\;
				
				$r.push(^{b}{\bf \hat{R}}_{o}^{j}-^{b}{\bf \hat{R}}_{o}^{k})$, $t.push({\bf P}_i^k-{\bf P}_i^j)$\;
				
				$R.push((^{b}{\bf \hat{R}}_{o}^{j})^{-1}-(^{b}{\bf \hat{R}}_{o}^{k})^{-1})$\;	
				
				$T.push((^{b}{\bf \hat{R}}_{o}^{j})^{-1} \times {\bf P}_i^j - (^{b}{\bf \hat{R}}_{o}^{k})^{-1} \times {\bf P}_i^k)$\;
            }
       }
       ${\bf \hat{q}}_i = pinv(r) \times t$ \label{step:eachq}\;
	   ${\bf \hat{Q}}_i = pinv(R) \times T$ \label{step:eachQ}\;					$RR.push(R)$, $TT.push(T)$\;	
    }
    ${\bf {\hat{Q}}} = pinv(RR) \times TT$ \label{step:overallQ}\;
    \tcp{Second: estimation of $^{IMU_b}\hat{\bf R}_{b}$}
	$a=[]$, $b=[]$\;
	\For{$i=1:6$}{ 
	   \For{$j=1:N_i$}{
	       $a.push(^{b}{\bf R}_{o}^{j} \times {\bf \hat{q}}_i)$\;
		   $b.push({\bf {\hat{Q}}} - {\bf P}_i)$\;
	   }
	}
    $\Delta_{^{IMU_b}\hat{\bf R}_{b}} = b \times pinv(a)$ \label{step:eachq}\;
    \tcp{compute overall errors}
    \For{$i=1:6$}{
       \For{$j=1:N_i-1$}{
            \For{$k=1+j:N_i$}{
                errors += $|(^{b}{\bf \hat{R}}_{o}^{j}-^{b}{\bf \hat{R}}_{o}^{k}) \times {\bf \hat{q}}_i + {\bf P}_i^j-{\bf P}_i^k|$\;
            }
            errors +=  $|{\bf {\hat{Q}}}_i - (\Delta_{^{IMU_b}\hat{\bf R}_{b}} \times ^{b}\hat{\bf R}_{o}^{j} \times {\bf \hat{q}}_i + {\bf P}_i^j)|$\;
         }
    }
}
${^{IMU_b}\hat{\bf R}_{b}} = \Delta_{^{IMU_b}\hat{\bf R}_{b}}$\;
\end{algorithm}

\subsubsection{Recap of calibration procedure}
\yhk{We summarize all calibration procedures in order as follows:}

\noindent
\yhk{
 {\noindent \bf Stereo PSDs calibration:}
\begin{tightenumerate}
	\item Collect 2D image points (more than 200 points over the whole workspace) using 4x3 checkerboard as shown Figure\,\ref{fig:calibration}. 
	\item The intrinsic/extrinsic and radial/tangential distortions parameters are optimized using the stereo camera calibration tool ({\em e.g.,}\,\citep{Bouguet01CameraCT}). 
	\item If a pincushion distortion exists, apply Equation\,\eqref{eq:distortion} at step 2)
	\item Finally, we can get ${\bf P}_{i}$.
\end{tightenumerate}
\noindent {\bf Calibration of IMU and PSD coordinate system:} The next step is to compute the tip position ${\bf Q}_i$ in Equation\,\eqref{eq:q}. There exist two parameters ($^{IMU_o}{\bf R}_{o}$, $^{IMU_b}{\bf R}_{b}$) associated with $^{b}{\bf R}_{o}$ that we have to calibrate.
\begin{tightenumerate}
	\item Collect samples (more than 5 points over possible unique orientation changes), which consists of one for the ground-truth orientation $^{g}{\bf \bar{R}}_{o}$ and another for TOU's pure IMU measure $^g{\bf R}_{IMU_o}$.
	\item Compute the overdetermined linear system by the pseudo-inverse (Equation\,\eqref{eq:object_tip_equation}), which provide the solution of $^{IMU_o}{\bf R}_{o}$ for $^{g}{\bf R}_{o}$ (Equation\,\eqref{eq:object}).
	\item Another collection of $N_i$ number of samples for $i$-th IR-LED, assuming all data have a common pivot point ${\bf Q}$. Samples are consisted of ${\bf P}_{i}$, $^g{\bf R}_{IMU_o}$, and $^g{\bf R}_{IMU_b}$.
	\item Construct two over determined linear equations; one for Equation\,\eqref{eq:estimationRo_process2}, another for Equation\,\eqref{eq:estimationRo2}
	\item Iteratively optimizing the solution of $^{IMU_b}{\bf R}_{b}$ until the threshold meet (Algorithm\,\ref{alg:iterative} explains 3-5 steps).
\end{tightenumerate}
}

\section{Experiment and Results}
\subsection{System setup}
\begin{figure}[t]
	\begin{center}
		\includegraphics[scale= 0.47]{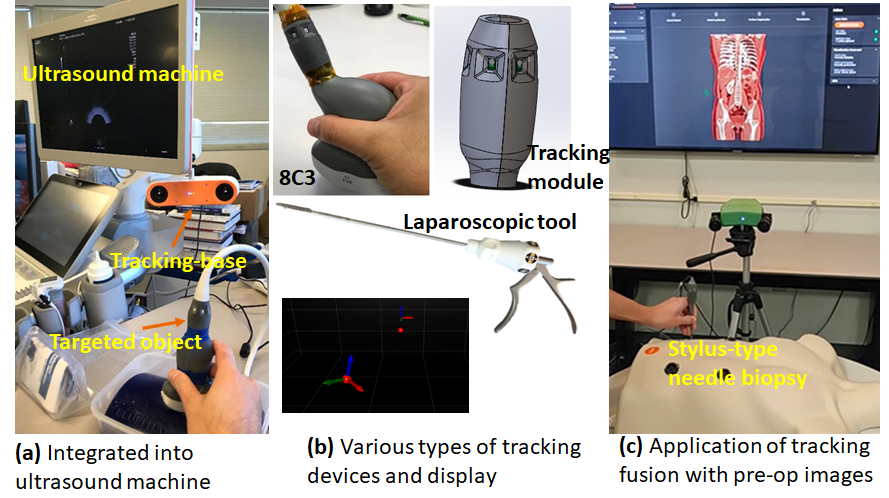}
		\caption{Our tracking module is a compact size, which can be attached to any rigid objects. (a) The tracking module is attached to the handle of the transducer, and the 6-DoF pose information fused with ultrasound images. (b) Our proposed system is also integrated with angled ultrasound probe, laparoscopic tool, and a stylus-type needle biopsy tool. The 6-DoF pose is displayed based on QT tools in real-time. (c) One exemplary demonstration of a pose tracking fusion with pre-op images using a stylus needle tool.
			\label{fig:system}}
		\vspace*{-10pt}
	\end{center}
\end{figure}

Figure\,\ref{fig:system} shows our proposed system  integrated and utilized within several applications ({\em e.g.,} ultrasound probes, laparoscopic tool, a stylus pen, a needle biopsy tool, etc.).
The TBU, with dimension (L x W x H) of $200 \times 50 \times 70~mm$, could be mounted to stationary frame while having a clear line-of-sight  to at least one IR-LEDs. The maximum frame rate is $100~Hz$, and the distance of IR-LEDs to tracking base could vary from  $300~mm$ to $1300~mm$. The object sensing dimension (L X W X H ~mm) is $50 \times 20 \times 20$, which depends on IR-LED ring size.

We use a UR5 robotic manipulator (Figure\,\ref{fig:calibration}) to calibrate and to measure the final tracking error for  performance analysis. UR5 is a 6-axis robot, which provides movement accuracy and repeatability of $\pm~0.1~mm$ in a working radius of $1000~mm$ with the pivot tool.

\yhk{We tested our proposed system in a typical lab setting in which we measured an interfering luminous flux, ranging between $40$ and $200$~lux due to sun light from windows, which can be transferred to the output voltage noise. We designed our proposed system to deal with the maximum $200$~lux, which could be canceled out by our approach (Section\,\ref{sec:cancel}). The signal-to-noise ratio (SNR) can be defined as the summation output signal level divided by the output voltage noise level, which is computed to be more than 30~db. In addition, the position resolution $PR$ is defined as a factor of the resistance length $L$ and $SNR$, $PR = L \times SNR$ (detailed in \,\citep{hamamatsu,thorlabs}). For this reason, the best position resolution can be achieved at $0.75~\mu m$ ({\em i.e.} for the output noise value of 300~$\mu V_{rms}$ with 4~V summation signal, which is controlled by our active light controllers), while the worst position resolution can in theory be $0.025~mm$  ({\em i.e.} when the output noise is 10~$mV_{rms}$ directly measured from ADC after cancellation for the signal summation value of 4~V).}



\subsection{Calibration setup}
{\noindent \bf Calibration of $^{IMU_o}{\bf R}_o$:~}
We designed a holder for TOU, which we mounted  to the end-effector of the UR5. 

UR5 provides the relative pose of the attached tool tip , i.e., tool center point (TCP) in the robot coordinate system. By attaching the tracked object aligned in lieu of a tool to the end-effector, we can get highly accurate pose of the tracked object, which we use as the ground truth.

To get $^{IMU_o}{\bf R}_o$, we collected 15 measurements  along each axis of the object, with a total number of  45 for x, y, z-axes. Each data point has a pair of $^{b}{\bf R}_o$ from UR5 as the ground-truth and $^{g}{\bf R}_{IMG_o}$ from IMU. Then, we used a least-square approximation method using a pseudo-inverse shown in Equation\,\eqref{eq:object_tip_equation}. The estimated $^{IMU_o}{\bf R}_o$ is defined as a quaternion, $(w ,x , y, z)=[0.9999, 0.0027, -0.0043, 0.0095]$. 

\vspace*{5pt}
{\noindent \bf Calibration of $^{IMU_b}{\bf R}_{b}$ and ${\bf q}_i$:~}

\begin{figure}[t]
	\begin{center}
		\includegraphics[scale= 0.4]{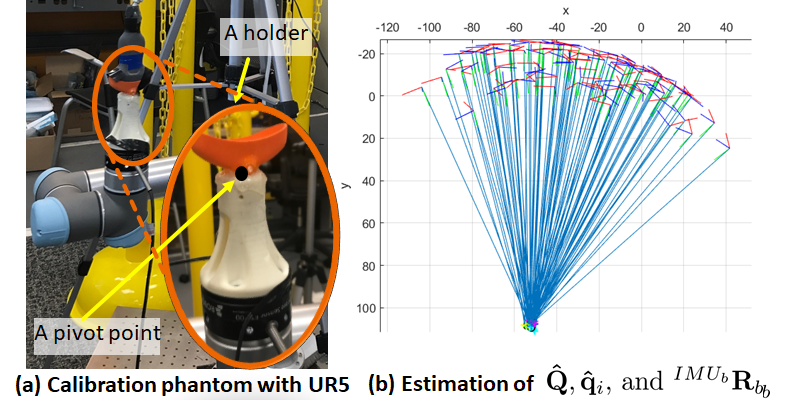}
		\vspace*{-5pt}
		\caption{ (a) A holder is attached to the end-effector of UR5, thus we can control the tool center point directly, which provides the ground truth for $^{b}{\bf R}_o$. (b) A set of ${\bf S}_i$ is plotted with position and orientation. Using the proposed iterative method that can estimate $^{IMU_b}{\bf R}_{b}$, ${\bf \hat{Q}}$, and ${\bf \hat{q}}_i$.  
			\label{fig:zig}}
	\end{center}
\end{figure}

Using UR5, we can freely rotate TOU about a pivot point (Figure\,\ref{fig:zig}(a)). We collected 15 measurements ${\bf S}_i$ from each of six IR-LEDs, rotating about a fixed pivot points. We move UR5 and change the pivot point to cover a large work space. Then we apply six set of ${\bf S}_i$ into Algorithm\,\ref{alg:iterative}.

The estimated $^{IMU_b}{\bf R}_{b}$ is written as a quaternion $(w, x, y, z)= [0.9967, -0.0201, -0.0713, -0.0323]$. The estimated ${\bf {Q}}$ is $(-52.59, 108.10, 780.70)$ while the standard deviation (unit=$~mm$) is $(0.79, 0.98, 1.02)$ among LEDs.

\subsection{Accuracy assessment and interpretation}

We analyze the accuracy of our proposed system in the following three scenarios;
\begin{tightenumerate}
	\item[1)] Static accuracy of 6-DoF pose over the three different size of area.
	\item[2)] Dynamic accuracy of 6-DoF pose at $100~mm$
	\item[3)] Power consumption and temperature over distance
\end{tightenumerate}

To evaluate the proposed method, we used Root-Mean-Square, Mean, and $95~\%$ confidence interval of the error, as it has been done for example for assessing OTS errors in\,\citep{wiles04accuracy}.

\begin{figure}[b]	
	\begin{center}
		\includegraphics[scale= 0.4]{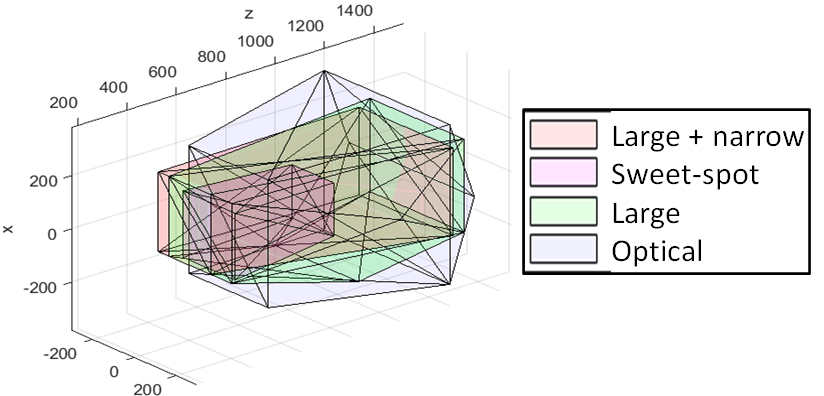}\label{fig:areas}
\vspace*{-10pt}			
	\caption{Overall boundary of workspaces that we evaluate are displayed. The orange area represents {\it the large, but narrow are}. The red area shows {\it the sweet-spot}. The green area shows {\it the large area}. The purple shows {\it OTS} (Vicra, NDI).
		\label{fig:curvature_errors}}
	\end{center}\vspace*{-15pt}
\end{figure}

\subsubsection{Static accuracy over the three different work space sizes}

To characterize the system accuracy dependency to work space size, we measured the error for three different hypothetical work spaces; (a) {\it the large area} with the volume of $350~\times~400~\times~900~mm$ ($158,500~cm^3$) with depth from $300~mm$ to $1300~mm$. (b) {\it the large + narrow area} with the volume of  $300~\times~300~\times~900~mm$ ($100,000~cm^3$), amd similar depth as large area {\it the large area}. (c) {\it the sweet-spot}, which is defined as $200~\times~300~\times~400~mm$ ($20,500~cm^3$), with the depth distance from $400~mm$ to $900~mm$.

The overall work space is shown in Figure\,\ref{fig:curvature_errors}, which additionally depicts an overlaid work space ($190,000~cm^3$ of OTS (Polaris Vicra, NDI\,\citep{ndi}).
To measure the static accuracy, the UR5 robot is programmed to cover the entire defined work spaces by a $10~cm$ interval for each dimension in 3D, while we collect data (${\bf \hat{Q}}$, $^b{\bf \hat{R}}_o$). 
\yhk{Over 400 samples are taken at each grid point. 958 grid points are used for {\it the large area}, 740 grid points are used for {\it the large + narrow area}, and 176 grid points are used for {\it the sweet-spot}.}
The ground truth of distance between any two samples can directly be computed from the actual Cartesian coordinates of UR5 end-effector. 


Table\,\ref{tab:static} shows the static position error; ${\bf \hat{Q}}$ computed by Equation\,\eqref{eq:q} including  $^b{\bf \hat{R}}_o$ and ${\bf P}_i$. As the static orientation $^b{\bf \hat{R}}_o$ error does not change over distance, we separately report the static accuracy of the orientation ($^b{\bf \hat{R}}_o$) at the last row. 
Overall, the position errors ($95\%$ CI) of our proposed system are computed to be $0.925~mm$, $1.934~mm$, and $2.254~mm$ for three different sizes of work space, respectively, and our orientation error of (0.059$^\circ$) is superior to OTS using  passive markers. Based on the report\,\citep{wiles04accuracy}, the static position error of Polaris Vicra is $0.462~mm$ ($95\%$ CI Table~2 of \citep{wiles04accuracy}), while the orientation errors is $0.713^\circ$ ($95\%$ CI) in an ideal setting.


\begin{table}[t]
	\centering
	\caption{Static accuracy and precision errors of PSD tracker, position/orientation error statistics}
	\label{tab:static}
	\scalebox{1}{\begin{tabular}{|l|c|c|c|c|}
		\hline
		         & \yhk{Precision} &  \multicolumn{3}{c|}{\begin{tabular}[c]{@{}c@{}}Accuracy errors \end{tabular}} \\\cline{3-5} 
		Regions  &  RMS			   & RMS  & Mean Error & $95\%$ CI \\
		\hline
		{\it The large}
		& \yhk{0.44~$mm$}
		& 1.369~$mm$     
		& 1.237~$mm$          
		& 2.254~$mm$      \\
		{\it The large+narrow}
		& \yhk{0.39~$mm$}    
		& 1.171~$mm$     
		& 0.846~$mm$         
		& 1.934~$mm$    \\
		{\it The sweet-spot} 
		& \yhk{0.22~$mm$}  
		& 0.56~$mm$     
		& 0.434~$mm$     
		& 0.925~$mm$    \\
		\hline\hline
		{\it 3D orientation} 
		& \yhk{0.03~$^{\circ}$}
		& 0.043$^{\circ}$     
		& 0.040$^{\circ}$     
		& 0.059$^{\circ}$  \\
		\hline
	\end{tabular}}
\end{table}

\subsubsection{Dynamic accuracy of 6-DoF pose at $100~cm$}

\begin{figure}[b!]	
	\begin{center}
			\includegraphics[scale= 0.38]{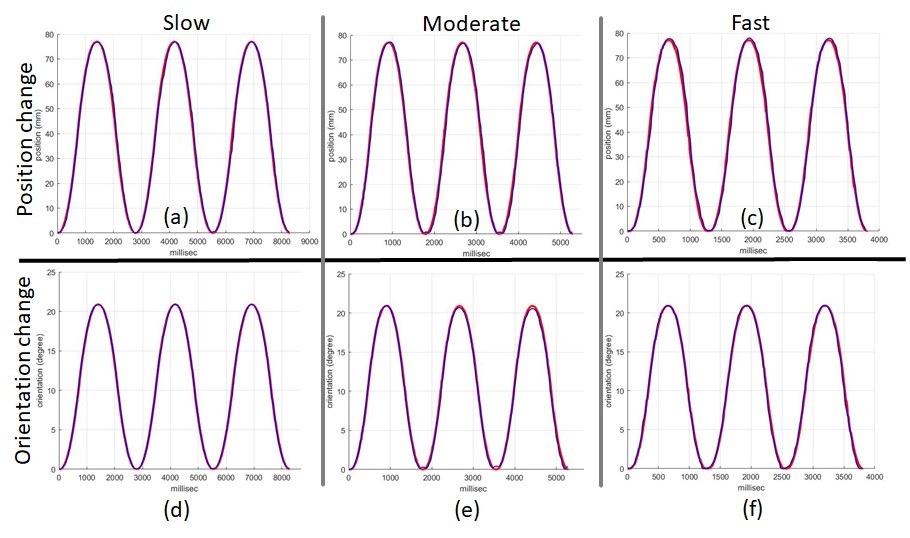}\label{fig:dynamic}	
	\end{center}\vspace*{-10pt}
	\caption{ Dynamic pose change of PSD tracker over time at $100~cm$. The position changes of three dynamic tests are (a), (b), and (c) while the orientation changes of three dynamic tests are (d), (e), and (f). The red line shows the ground-truth from UR5 API for actual Cartesian coordinates of the tool. The blue line shows pose changes of our tracker.
		\label{fig:dynamic_accuracy}}
\end{figure}

To show the dynamic accuracy of the proposed approach, we moved TOU following a pre-defined sway motion using the UR5, which repeatedly moves TOU by changing its translation and rotation simultaneously within range of $75~mm$,  $20^\circ$, respectively. We created three cycle of this trajectory with three different velocity and acceleration; (1) {\it Slow}  $7~cm/sec$ and $20~cm/sec^2$, (2) {\it Moderate}  $14~cm/sec$ and $50~cm/sec^2$, (3) {\it Fast}  $25~cm/sec$ and $100~cm/sec^2$. The test distance between TBU and TOU is about $100~cm$.
We collected the position ${\bf \hat{Q}}$ and the orientation $^b{\bf R}_o$, for the system, while the ground-truth is acquired from UR5 application programming interface (API) in real-time. Figure\,\ref{fig:dynamic_accuracy} depicts the overlaid pose changes over time.

\begin{table}[t]
	\centering
	\caption{Dynamic pose accuracy of PSD tracker, distance/orientation error}
	\label{tab:dynamic}
	\scalebox{1.1}{\begin{tabular}{|l|c|c|c|}
			\hline
			Condition  & RMS   & Mean Error  & $95\%$ CI  \\ 
			           & ($mm$, degree~$^\circ$)   & ($mm$, degree~$^\circ$)  & ($mm$, degree~$^\circ$)\\ 
			\hline
			{\it Slow }		 
			& (0.743, 0.135)     
			& (0.725, 0.159)          
			& (1.222, 0.222)      \\
			\hline
			{\it Moderate }
			& (0.825, 0.246)     
			& (0.988, 0.327)         
			& (1.357, 0.405)    \\
			\hline
			{\it Fast }     
			& (0.906, 0.323)     
			& (1.052, 0.413)         
			& (1.49, 0.532)    \\
			\hline
	\end{tabular}}
\end{table}

The overall dynamic performance evaluation is described in Table\,\ref{tab:dynamic}. The orientation errors are within sub-millimeter range for all testing conditions. The dynamic position error increases as compared to static one by less than a millimeter. Unfortunately, dynamic measurements error are not provided in many related work making it hard to compare. The results overall demonstrates that the proposed system can be a competitive system in terms of accuracy for tracking  rigid tools to other commercially available OTS, ETS, and video-metric based tracking systems.

\subsubsection{Power and temperature measurement}

\begin{figure}[t!]	
	\begin{center}
		\includegraphics[scale= 0.4]{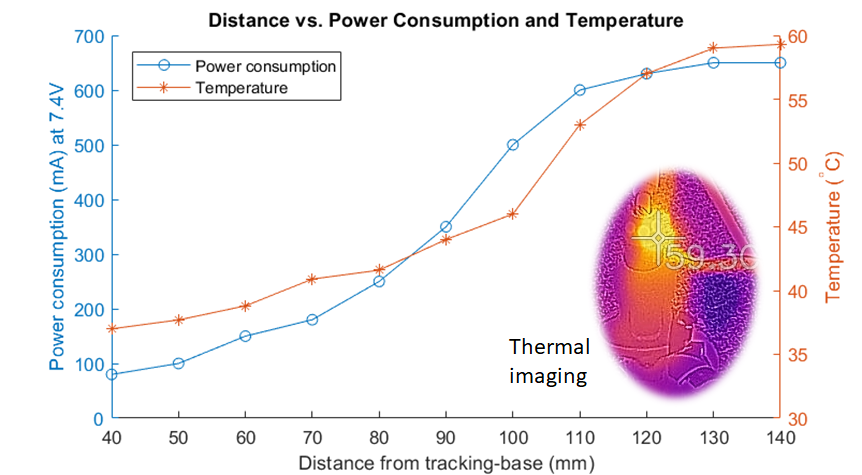}
		\caption{\yhk{Demonstration of power consumption and temperature changes over distance: X-axis is distance between TBU and TOU. Left Y-axis is power consumption (mA) at $7.4~V$. Right Y-axis is temperature ($^\circ C$). We used FLIR one device to measure temperature, power consumption is directly measured via the digital power supply. All values are collected after saturated at each point. Each value measures how much our proposed system spends power and radiates heat accordingly over the whole work space.
		}
			\label{fig:power_temperature}}
	\end{center}\vspace*{-10pt}
\end{figure}

We evaluated the power consumption and the temperature of the unit during nominal operation. TOU is moved in a straight line from $40~cm$ to $140~cm$ away from TBU. \yhk{We used the FLIR One thermal imaging sensor (Teledyne FLIR LLC, Oregan, USA) to measure the surface temperature while power consumption is directly measured by an digital power supply. TOU was kept stationary for one minute at each point along the path, then we gathered saturated temperature and power consumption as shown in Figure\,\ref{fig:power_temperature}.} The supplied voltage for TOU was $7.4~V$, and the current and temperature at the minimum distance ($40~cm$) was $90~mA$ and $37^\circ C$, respectively, whereas the same at the maximum distance of  ($140~cm$) were measured to be $650~mA$ and $59^\circ C$.



\section{Discussion}

%

%

\begin{tightitemize}
	\item We demonstrated our proposed pattern-based IR-LED identification method with active power controls performs well in terms of both accuracy and power consumption. We believe that there is a possibility to increase tracking performance by decreasing PWM duty cycle, which could translate to higher IR-LED voltage and increase LED luminance and overall improved accuracy.  Moreover, with the reduced PWM duty cycle, we could also decrease the overall power consumption leading to decreased temperature. Our current design has multiple LEDs in TOU, and in order to further optimize the power consumption, the specific IR-LEDs with clear light of sight to the tracking base could be illuminated. This could be possible by analyzing the relative orientation of TOU with respect to TBU in real-time, and changing the LED firing pattern accordingly. Overall, these several additional options could contribute to further miniaturization of the sensing unit with a light weight module supplied by the small size lithium ion battery.	
	\item Our proposed system also requires the line-of-sight condition similiar to OTS. However, our system has multiple IR-LEDs covering $360^\circ$, therefore, if we could potentially use multiple TBUs so that one tracking base system can at least track one IR-LED. In this scenario, all TBU stations can be calibrated at once in the working area. Moreover, the base coordinate system can be transferred to the any fixed reference frame of one of the TBU systems in the same space by using Equation\,\eqref{eq:base}.
	\item Pattern-based identification method could not track a large number of TOUs due to a bandwidth limitation we have due to minimum requirement on PWM duty cycle used within the IR-LED identification pattern. However, this problem could potentially be addressed by using time division multiplexing, frequency multiplexing, or a combination. These topics are beyond the scope of this paper, and could be considered as an extension of the work. 
\end{tightitemize}

\section{Conclusions}
We proposed a novel 6-DoF pose tracking system that uses stereo-based PSDs and multiple IMUs, which are both cost efficient components widely used in many other applications. We devised a practical IR-LED identification methodology with efficient power control to provide tracking accuracy within a large tracking work space. High refresh rates for both PSDs and IMUs provided an opportunity for the overall 6-DoF pose tracking system also to have a high update rate and low latency. Furthermore, the proposed tracking sensors could be manufactured into a small form factor, which makes it favourable for a variety of applications. 

\yhk{Results demonstrate that the static position RMS error is between $0.6~mm$ at the sweet spot ({\em i.e.}, mid range of the work space) and $1.3~mm$ at the largest extend of the work space, while the precision error is $0.22~mm$ and $0.44~mm$, respectively. The dynamic position RMS error at $100~mm$ is between $0.7~mm$ at a slow motion (7~cm/s) and $0.9~mm$ at a fast motion (25~cm/s). The orientation RMS error is $0.04^\circ$ at static and $0.9^\circ$ at a fast dynamic motion. We have also measured both the consumed power and the temperature (at the rigid object side) as functions of distance between tracking base and object, which demonstrate that our miniature sized ($50 \times 20 \times 20~(mm)$) tracked-object unit temperature gets saturated between $37^\circ$C and $58^\circ$C over the whole work space.}

Overall, the results demonstrated that the proposed tracking system can be used as a  wide-area, low-latency, and power-efficient 6-DoF pose tracking system as an alternative inexpensive option to OTS and EMT.

\section*{Disclaimer}
	The concepts and information presented in this paper are based on research results that are not commercially available. Future availability cannot be guaranteed.

{
	\bibliography{references_sensors}
}

\end{document}